\documentclass[conference]{IEEEtran}

\usepackage[utf8]{inputenc}
\usepackage{graphicx}
\usepackage{float}
\usepackage{multirow}  
\usepackage{amssymb}
\usepackage{amsmath}

\usepackage[colorlinks=true, citecolor=blue,linkcolor=blue,bookmarks=false]{hyperref}
\usepackage[labelsep=quad,indention=10pt]{subfig}

\newcommand{\loss}{\text{loss}}

\usepackage{adjustbox}
\usepackage[style=ieee]{biblatex} 
\bibliography{references}

\title{ATRAS: Adversarially Trained Robust Architecture Search}

\author{
\IEEEauthorblockN{Yigit Alparslan}
\IEEEauthorblockA{\textit{Department of Computer Science} \\
\textit{Drexel University}\\
Philadelphia, PA, US \\
ya332@drexel.edu}
\and
\IEEEauthorblockN{Edward Kim}
\IEEEauthorblockA{\textit{Department of Computer Science} \\
\textit{Drexel University}\\
Philadelphia, PA, US \\
ek826@drexel.edu}
}

\begin{document}

\maketitle

\begin{abstract}
In this paper, we explore the effect of architecture completeness on adversarial robustness. We train models with different architectures on CIFAR-10 and MNIST dataset. For each model, we vary different number of layers and different number of nodes in the layer. For every architecture candidate, we use Fast Gradient Sign Method (FGSM) to generate untargeted adversarial attacks and use adversarial training to defend against those attacks. For each architecture candidate, we report pre-attack, post-attack and post-defense accuracy for the model as well as the architecture parameters and the impact of completeness to the model accuracies.

\end{abstract}


\let\thefootnote\relax\footnote{{All code is open-sourced on \href{https://github.com/ya332/ATRAS}{GitHub.}}}

\section{Introduction}
\label{sec:introduction}
Recent machine learning breakthroughs help solve many tasks including facial recognition \cite{aajfacialadversarial}, surveillance \cite{aplkc20}, natural language processing tasks \cite{daelemans2002evaluation}, materials discovery \cite{edwardkimmaterialsdiscovery} and bio-authentication systems \cite{deng2019arcface}. Yet, existence of adversarial attacks in deep learning models trouble many researchers since mission-critical systems where these models are deployed have very little room for errors which might be caused by adversarial attacks \cite{alparslan2021evaluating} \cite{alparslan2021robust}. 
\\
In this study, we are interested in exploring what steps we can take to improve robustness from the defender's point of view as well as what steps we can take to adversarially attack a model more efficiently from the attacker's point of view. In both attempts, our goal is to understand adversarial robustness better and propose a framework to increase robustness.

The title of this study, ``ATRAS'', is an acronym for "Adversarially Trained Robust Architecture Search". 

We use adversarial training to defend our deep learning models since adversarial training is studied in the literature and remains to be a strong defense at the time of the writing of this paper \cite{shafahi2019adversarial} \cite{wang2019convergence} \cite{tramèr2020ensemble} \cite{alparslan2021robust}. We investigate how architecture size impacts amount of accuracy that is recovered after adversarially training a model against attacks. Besides adversarial training, we also investigate how architecture size impacts transferability of adversarial attacks.\\
\\
This research is organized so that \autoref{sec:introduction} introduces the concept of drowsy driving and \autoref{sec:relatedwork} explores what has been done in this field. \autoref{sec:methodology} explains the dataset, the models, adversarial attacks, adversarial training and limitations that we use/have in this study. We report the results in \autoref{sec:results} and conclude the study in \autoref{sec:conclusion} with summarizing what we have done in this study and discussing where the research might go in the future.

\section{Related Work}
\label{sec:relatedwork}

Adversarial attacks are inputs that look like the original images but with perturbations added to result in misclassifications in the classifier \cite{AdvAttacks} \cite{kim2020modeling} \cite{edwardkimregularization} \cite{kim2018deep}. Adversarial attacks can be created in the image domain \cite{aajfacialadversarial} as well as audio domain \cite{aab20audio}.

Adversarial training is one way of defending against these attacks since using adversarial attacks \cite{madry2017towards}. we can generate adversarial samples and then use these samples in our training to develop high accuracy models. Adversarial training as a defense depends on model and task at hand significantly.

\section{Methodology}
\label{sec:methodology}

\subsection{Adversarial Attacks}
In this study, we only use FGSM because only FGSM out of DeepFool and PGD runs reasonably fast on a commodity computer to finish all the desired workloads.

\subsubsection{Fast Gradient Sign}
\label{attack:fgs}
The fast gradient sign \cite{goodfellow2014explaining} method optimizes for the $L_{\infty}$ distance metric and its advantage is fast running tim, which comes at the expense of generating images that are very similar to the original image.

Given an image $x$ the fast gradient sign method sets
\begin{equation*}
  x' = x - \epsilon \cdot \text{sign}(\nabla \loss_{F,t}(x)),
\end{equation*}
where $\epsilon$ is chosen to be sufficiently small so as to be undetectable,
and $t$ is the target label. 
Intuitively, for each pixel, the fast gradient sign method uses the gradient
of the loss function to determine in which direction the pixel's intensity should be changed (whether it should be increased or decreased) to minimize the loss function; then, it shifts all pixels
simultaneously.

\subsection{Adversarial Training}
Adversarial training \cite{fgm} \cite{kannan2018adversarial} \cite{madry2017towards} is the process where the adversarially generated samples are included in the training data in the hopes that the model will recognize the attacks next time sees it. In the current literature, adversarial training is one of the stronger defenses against adversarial attacks especially if it is combined with other defenses \cite{madry2017towards} \cite{tramer2020adaptive}. We adversarially train our MNIST and CIFAR-10 models and  report  pre attack, post attack and post defense train and test accuracies as it can be seen in ~\autoref{table:fgsm_cifar10} and  ~\autoref{table:fgsm_mnist}. During our adversarial training, we can only use Fast Gradient Sign Method due to its very fast execution compared to other computationally expensive attacks such as DeepFool and Projected Gradient Sign Method.




\subsection{Dataset}

In this study, we use CIFAR-10 and MNIST datasets.

\begin{figure}
\centering
  \includegraphics[width=0.9\columnwidth]{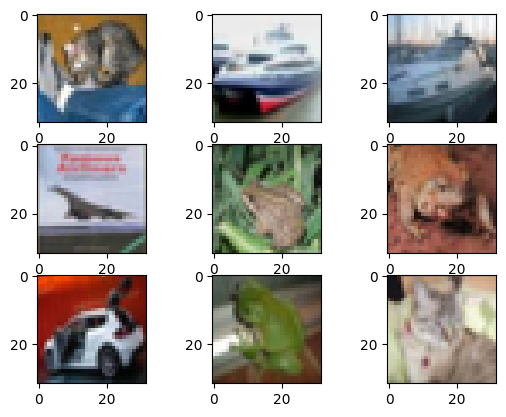}
  \caption{Samples from CIFAR-10 dataset}~\label{fig:cew_dataset}
\end{figure}

\begin{figure}
\centering
  \includegraphics[width=0.9\columnwidth]{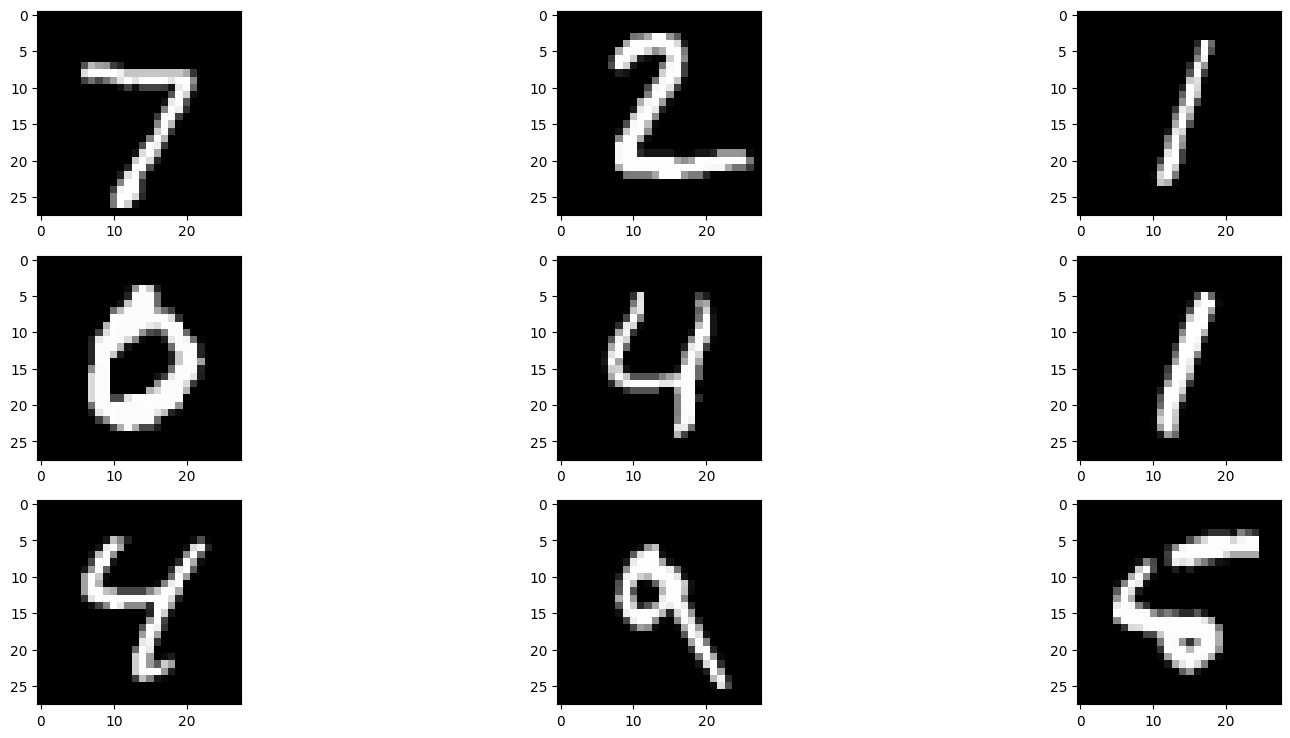}
  \caption{Samples from MNISt dataset. }~\label{fig:figure2}
\end{figure}

\subsection{Limitations}
In this study, we examine many different combinations to see the full effect of attacks and defenses. Due to numerous workloads required to run in this research paper, we were able to use only FGSM due to its fast execution. This limited our research to only run FGSM as adversarial attack. Another limitation that arised was the dataset. Running many workloads over and over again required too much computational resource so we were only able to use CIFAR-10 and MNIST datasets and had to exclude ImageNet dataset.\\

\section{Experiment Results and Evaluation}
\label{sec:results}

We report all the results in \autoref{table:fgsm_mnist} and \autoref{table:fgsm_cifar10}. In our analysis, we need to consider two aspects of the attacks. 
\begin{itemize}
    \item How successful is the adversarial attack? 
    \item How successful is the defense?
\end{itemize}

To answer these questions, we adversarially attack the models before they are adversarially trained. Later, we use adversarial training to train the models and defend them against adversarial attacks. We then adversarially attack the models after they are adversarially trained. The increase in accuracies between these two rounds of attacks is important because it tells us how much the model can recover (i.e the accuracy is previously low due to attacks. Now, the model is trained adversarially, how much accuracy can we recover?) This "recovery" measure is therefore important to answer how successful the adversarial attack is as well as how successful the defense is.
\\
\\
For MNIST, we see that adversarial training as a defense is most effective on smaller architectures. Below, we report differences between accuracies \autoref{table:fgsm_mnist}, model candidates with when attacked after adversarial training compared to when attacked before adversarial training. 

\begin{enumerate}
    \item Models with one hidden layer average a 4.58\% increase
    \item Models with two hidden layers average 2.5\% decrease
    \item Models with three hidden layers average 4.06\% increase
    \item Models with any other larger architecture average 0.99\% increase
\end{enumerate}
  when attacked after adversarial training compared to when attacked before adversarial training. 
 \\
 \\
For CIFAR-10, we see that adversarial training as a defense is most effective on smaller architectures. Below, we report differences between accuracies \autoref{table:fgsm_mnist}, model candidates with when attacked after adversarial training compared to when attacked before adversarial training. 

\begin{enumerate}
    \item Models with one hidden layer average a 35.2\% increase
    \item Models with two hidden layers average 44.5\% decrease
    \item Models with three hidden layers average 31.30\% increase
    \item Models with any other larger architecture average 7.28\% increase
\end{enumerate}
  when attacked after adversarial training compared to when attacked before adversarial training. For CIFAR-10, we also report that post defense accuracy against adversarial attacks is the largest among smaller networks (around 53\% test accuracy compared to 43.34\% test accuracy) 
  
  We highlight the following observations before we conclude the paper:
  
  \begin{enumerate}
      \item Models with smaller architectures take shorter to train and therefore, become preferable for adversarial training due to less time that they take.
      \item Post defense accuracy when adversarially attacked is the highest on models with smaller architectures 
      ]\item Most recovery is observed on models with smaller architectures (4.58\% recovery increase in MNIST and  35.2\% recovery increase in CIFAR-10, both highest compared to all other models)
      
  \end{enumerate}
  
  We highlight again that from defender's point of view, both of these findings suggest that smaller networks are more efficient to use since they have less parameters so it is faster to adversarially train them and smaller networks can recover their pre-attack accuracies almost all the time. From the attacker's point of view, smaller networks are a gateway to attack larger networks since adversarial attacks on a smaller network are also adversarial to larger networks due to transferability principle.

\begin{table*}[h]
\caption{FGSM attacks on MNIST. We report a base line training and test accuracy. We then generate adversarial attack success rate before the defense. We then use adversarial training to defend our model against adversarial attacks and report the accuracy when attacked after the defense}
\label{table:fgsm_mnist}
\begin{tabular}{|l|l|l|l|l|l|l|}
\hline
\textbf{train\_acc} & \textbf{test\_acc} & \textbf{\parbox[t]{2.5cm}{acc\_when\_attacked \\ \_before\_adv\_training}} & \textbf{adversarial\_train\_acc} & \textbf{adversarial\_test\_acc} & \textbf{\parbox[t]{2.5cm}{acc\_when\_attacked \\ \_after\_adv\_training}} & \textbf{hidden\_layers}                  \\ \hline
0.1126                   & 0.1142                  & 0.1142                                              & 0.1126                                & 0.1142                               & 0.1142                                             & {[}8, 16, 32, 64, 128, 256, 512, 1024{]} \\ \hline
0.762                    & 0.7274                  & 0.6864                                              & 0.9886                                & 0.959                                & 0.9546                                             & {[}16, 32, 64, 128, 256, 512, 1024{]}    \\ \hline
0.1126                   & 0.1142                  & 0.1142                                              & 0.1126                                & 0.1142                               & 0.1142                                             & {[}24, 48, 96, 192, 384, 768{]}          \\ \hline
0.98                     & 0.9544                  & 0.9238                                              & 0.9918                                & 0.9644                               & 0.9616                                             & {[}32, 64, 128, 256, 512, 1024{]}        \\ \hline
0.11                     & 0.1024                  & 0.1024                                              & 0.1126                                & 0.1142                               & 0.1142                                             & {[}40, 80, 160, 320, 640{]}              \\ \hline
0.1126                   & 0.1142                  & 0.1142                                              & 0.1126                                & 0.1142                               & 0.1142                                             & {[}48, 96, 192, 384, 768{]}              \\ \hline
0.1126                   & 0.1142                  & 0.1142                                              & 0.1126                                & 0.1142                               & 0.1142                                             & {[}56, 112, 224, 448, 896{]}             \\ \hline
0.8784                   & 0.8416                  & 0.727                                               & 0.9957                                & 0.9674                               & 0.9682                                             & {[}64, 128, 256, 512, 1024{]}            \\ \hline
0.992                    & 0.9672                  & 0.9418                                              & 0.9985                                & 0.9732                               & 0.9678                                             & {[}8, 32, 128, 512{]}                    \\ \hline
0.9896                   & 0.9682                  & 0.9216                                              & 0.9995                                & 0.9802                               & 0.9786                                             & {[}16, 64, 256, 1024{]}                  \\ \hline
0.9934                   & 0.9698                  & 0.9254                                              & 0.9992                                & 0.9754                               & 0.9754                                             & {[}24, 96, 384{]}                        \\ \hline
0.9896                   & 0.9586                  & 0.9154                                              & 0.9993                                & 0.9746                               & 0.9742                                             & {[}32, 128, 512{]}                       \\ \hline
0.9944                   & 0.972                   & 0.9462                                              & 0.9977                                & 0.9714                               & 0.9712                                             & {[}40, 160, 640{]}                       \\ \hline
0.9968                   & 0.9734                  & 0.9506                                              & 0.9984                                & 0.9772                               & 0.9762                                             & {[}48, 192, 768{]}                       \\ \hline
0.9954                   & 0.9738                  & 0.9426                                              & 0.9988                                & 0.975                                & 0.9776                                             & {[}56, 224, 896{]}                       \\ \hline
0.9964                   & 0.9728                  & 0.938                                               & 0.9989                                & 0.9788                               & 0.9806                                             & {[}64, 256, 1024{]}                      \\ \hline
0.995                    & 0.9654                  & 0.934                                               & 0.9994                                & 0.9726                               & 0.9712                                             & {[}8, 64, 512{]}                         \\ \hline
0.9938                   & 0.969                   & 0.9278                                              & 0.9991                                & 0.9778                               & 0.9788                                             & {[}16, 128, 1024{]}                      \\ \hline
0.9956                   & 0.9688                  & 0.9274                                              & 0.9994                                & 0.9738                               & 0.975                                              & {[}24, 192{]}                            \\ \hline
0.9966                   & 0.9678                  & 0.9398                                              & 0.9997                                & 0.9718                               & 0.97                                               & {[}32, 256{]}                            \\ \hline
0.9982                   & 0.9706                  & 0.948                                               & 1.0                                   & 0.9768                               & 0.9758                                             & {[}40, 320{]}                            \\ \hline
0.9954                   & 0.9716                  & 0.9446                                              & 0.9993                                & 0.9736                               & 0.9764                                             & {[}48, 384{]}                            \\ \hline
0.9968                   & 0.9704                  & 0.9436                                              & 0.9998                                & 0.9766                               & 0.9758                                             & {[}56, 448{]}                            \\ \hline
0.998                    & 0.9726                  & 0.9436                                              & 0.9998                                & 0.973                                & 0.9748                                             & {[}64, 512{]}                            \\ \hline
0.996                    & 0.972                   & 0.9272                                              & 0.9998                                & 0.9772                               & 0.9714                                             & {[}8, 128{]}                             \\ \hline
0.9946                   & 0.9696                  & 0.9246                                              & 0.9997                                & 0.9778                               & 0.9772                                             & {[}16, 256{]}                            \\ \hline
0.9982                   & 0.97                    & 0.9416                                              & 0.9998                                & 0.977                                & 0.9718                                             & {[}24, 384{]}                            \\ \hline
0.9976                   & 0.9726                  & 0.9358                                              & 0.9983                                & 0.964                                & 0.9704                                             & {[}32, 512{]}                            \\ \hline
0.9976                   & 0.9704                  & 0.9440                                  & 0.9996                                & 0.9716                               & 0.9734                                             & {[}40, 640{]}                            \\ \hline
0.9974                   & 0.9762                  & 0.9494                                              & 0.9997                                & 0.9722                               & 0.9782                                             & {[}48, 768{]}                            \\ \hline
0.9972                   & 0.9684                  & 0.9470                                  & 0.9999                                & 0.9774                               & 0.9762                                             & {[}56, 896{]}                            \\ \hline
0.9956                   & 0.9706                  & 0.9374                                              & 0.9995                                & 0.9744                               & 0.976                                              & {[}64, 1024{]}                           \\ \hline
0.9958                   & 0.9678                  & 0.9382                                              & 0.9992                                & 0.9748                               & 0.966                                              & {[}8, 256{]}                             \\ \hline
0.9966                   & 0.9686                  & 0.9384                                              & 0.9993                                & 0.9762                               & 0.9764                                             & {[}16, 512{]}                            \\ \hline
0.9978                   & 0.9766                  & 0.9472                                              & 0.9998                                & 0.9742                               & 0.9754                                             & {[}24, 768{]}                            \\ \hline
0.9984                   & 0.9742                  & 0.9438                                              & 0.9998                                & 0.9762                               & 0.9778                                             & {[}32, 1024{]}                           \\ \hline
0.9982                   & 0.9606                  & 0.9176                                              & 0.9999                                & 0.9712                               & 0.9704                                             & {[}40{]}                                 \\ \hline
0.9986                   & 0.9658                  & 0.9234                                              & 1.0                                   & 0.9714                               & 0.9692                                             & {[}48{]}                                 \\ \hline
0.9984                   & 0.9614                  & 0.924                                               & 1.0                                   & 0.9688                               & 0.9656                                             & {[}56{]}                                 \\ \hline
0.9988                   & 0.9646                  & 0.925                                               & 0.9999                                & 0.9654                               & 0.968                                              & {[}64{]}                                 \\ \hline
\end{tabular}
\end{table*}

\begin{table*}[h]
\caption{FGSM attacks on CIFAR10. We report a base line trainng and test accuracy. We then generate adversarial attack success rate before the defense. We then use adversarial training to defend our model againsts adversarial attacks and report the accuracy when attacked after the defense}
\label{table:fgsm_cifar10}
\begin{tabular}{|l|l|l|l|l|l|l|}
\hline
\textbf{train\_acc} & \textbf{test\_acc} & \textbf{\parbox[t]{2.5cm}{acc\_when\_attacked \\ \_before\_adv\_training}} & \textbf{adversarial\_train\_acc} & \textbf{adversarial\_test\_acc} & \textbf{\parbox[t]{2.5cm}{acc\_when\_attacked \\ \_after\_adv\_training}} & \textbf{hidden\_layers}                  \\ \hline
0.1040      & 0.1008                  & 0.1008                                              & 0.1038                                & 0.0982                               & 0.0982                                             & {[}8, 16, 32, 64, 128, 256, 512, 1024{]} \\ \hline
0.1040      & 0.1008                  & 0.1008                                              & 0.1040                   & 0.1008                               & 0.1008                                             & {[}16, 32, 64, 128, 256, 512, 1024{]}    \\ \hline
0.1038                   & 0.1024                  & 0.1024                                              & 0.1040                   & 0.1008                               & 0.1008                                             & {[}24, 48, 96, 192, 384, 768{]}          \\ \hline
0.1038                   & 0.1014                  & 0.1014                                              & 0.1038                                & 0.0982                               & 0.0982                                             & {[}32, 64, 128, 256, 512, 1024{]}        \\ \hline
0.1038                   & 0.1014                  & 0.1014                                              & 0.1038                                & 0.1014                               & 0.1014                                             & {[}40, 80, 160, 320, 640{]}              \\ \hline
0.413                    & 0.386                   & 0.1742                                              & 0.8623                                & 0.5614                               & 0.5386                                             & {[}48, 96, 192, 384, 768{]}              \\ \hline
0.1038                   & 0.0982                  & 0.0982                                              & 0.1038                                & 0.1024                               & 0.1024                                             & {[}56, 112, 224, 448, 896{]}             \\ \hline
0.1038                   & 0.1014                  & 0.1014                                              & 0.1038                                & 0.1014                               & 0.1014                                             & {[}64, 128, 256, 512, 1024{]}            \\ \hline
0.3938                   & 0.3862                  & 0.1766                                              & 0.6851                                & 0.5224                               & 0.4938                                             & {[}8, 32, 128, 512{]}                    \\ \hline
0.1038                   & 0.0982                  & 0.0982                                              & 0.6105                                & 0.5172                               & 0.5104                                             & {[}16, 64, 256, 1024{]}                  \\ \hline
0.5694                   & 0.4964                  & 0.1772                                              & 0.9328                                & 0.5720                 & 0.541                                              & {[}24, 96, 384{]}                        \\ \hline
0.5246                   & 0.4826                  & 0.1886                                              & 0.9277                                & 0.5856                               & 0.5524                                             & {[}32, 128, 512{]}                       \\ \hline
0.5262                   & 0.4820     & 0.1810                                 & 0.9365                                & 0.5898                               & 0.5606                                             & {[}40, 160, 640{]}                       \\ \hline
0.1038                   & 0.1024                  & 0.1024                                              & 0.1040                   & 0.1008                               & 0.1008                                             & {[}48, 192, 768{]}                       \\ \hline
0.568                    & 0.4984                  & 0.1794                                              & 0.9653                                & 0.5932                               & 0.56                                               & {[}56, 224, 896{]}                       \\ \hline
0.535                    & 0.4679     & 0.1939                                 & 0.9513                                & 0.5748                               & 0.5488                                             & {[}64, 256, 1024{]}                      \\ \hline
0.4970       & 0.4542                  & 0.1816                                              & 0.8340                    & 0.524                                & 0.4992                                             & {[}8, 64, 512{]}                         \\ \hline
0.5364                   & 0.4932                  & 0.192                                               & 0.9446                                & 0.5684                               & 0.5444                                             & {[}16, 128, 1024{]}                      \\ \hline
0.718                    & 0.5482                  & 0.1902                                              & 0.9936                                & 0.5492                               & 0.4954                                             & {[}24, 192{]}                            \\ \hline
0.718                    & 0.5498                  & 0.1984                                              & 0.9887                                & 0.5718                               & 0.5326                                             & {[}32, 256{]}                            \\ \hline
0.6968                   & 0.5398                  & 0.2034                                              & 0.9753                                & 0.5674                               & 0.528                                              & {[}40, 320{]}                            \\ \hline
0.664                    & 0.547                   & 0.1752                                              & 0.9763                                & 0.579                                & 0.526                                              & {[}48, 384{]}                            \\ \hline
0.659                    & 0.5332                  & 0.2022                                              & 0.9735                                & 0.5682                               & 0.5324                                             & {[}56, 448{]}                            \\ \hline
0.7002                   & 0.5608                  & 0.1936                                              & 0.9745                                & 0.5760                   & 0.5326                                             & {[}64, 512{]}                            \\ \hline
0.6676                   & 0.5344                  & 0.1624                                              & 0.9534                                & 0.5372                               & 0.4982                                             & {[}8, 128{]}                             \\ \hline
0.6714                   & 0.527                   & 0.196                                               & 0.9679                                & 0.562                                & 0.5408                                             & {[}16, 256{]}                            \\ \hline
0.6504                   & 0.5464                  & 0.185                                               & 0.9306                                & 0.573                                & 0.5124                                             & {[}24, 384{]}                            \\ \hline
0.6522                   & 0.5468                  & 0.1924                                              & 0.9452                                & 0.5764                               & 0.529                                              & {[}32, 512{]}                            \\ \hline
0.6954                   & 0.5334                  & 0.1992                                              & 0.9832                                & 0.5816                               & 0.5422                                             & {[}40, 640{]}                            \\ \hline
0.7138                   & 0.5589      & 0.1914                                              & 0.9888                                & 0.5824                               & 0.5328                                             & {[}48, 768{]}                            \\ \hline
0.748                    & 0.55399      & 0.2068                                              & 0.9955                                & 0.5844                               & 0.5446                                             & {[}56, 896{]}                            \\ \hline
0.6774                   & 0.5354                  & 0.1898                                              & 0.983                                 & 0.5636                               & 0.5308                                             & {[}64, 1024{]}                           \\ \hline
0.6                      & 0.5022                  & 0.1746                                              & 0.9217                                & 0.5466                               & 0.4974                                             & {[}8, 256{]}                             \\ \hline
0.6264                   & 0.5158                  & 0.1932                                              & 0.937                                 & 0.5692                               & 0.5198                                             & {[}16, 512{]}                            \\ \hline
0.6802                   & 0.5574                  & 0.1634                                              & 0.977                                 & 0.5714                               & 0.516                                              & {[}24, 768{]}                            \\ \hline
0.684                    & 0.5542                  & 0.196                                               & 0.9865                                & 0.5728                               & 0.539                                              & {[}32, 1024{]}                           \\ \hline
0.8608                   & 0.5726                  & 0.1644                                              & 0.9958                                & 0.5238                               & 0.5258                                             & {[}40{]}                                 \\ \hline
0.9036                   & 0.5596                  & 0.189                                               & 0.9983                                & 0.5314                               & 0.521                                              & {[}48{]}                                 \\ \hline
0.9079       & 0.5564                  & 0.184                                               & 0.9985                                & 0.5486                               & 0.5446                                             & {[}56{]}                                 \\ \hline
0.887                    & 0.5666                  & 0.179                                               & 0.9961                                & 0.54                                 & 0.5328                                             & {[}64{]}                                 \\ \hline
\end{tabular}
\end{table*}

\section{Conclusion and Future Work}
\label{sec:conclusion}
In this study, we attack models with different architecture and report that the attacks on a smaller network are also adversarial to larger networks. We use adversarial training to defend against adversarial attacks and see that smaller networks can recover more compared to larger networks. From defender's point of view, both of these findings suggest that smaller networks are more efficient to use since they have less parameters so it is faster to adversarially train them and smaller networks can recover their pre-attack accuracies almost all the time. From the attacker's point of view, smaller networks are a gateway to attack larger networks since adversarial attacks on a smaller network are also adversarial to larger networks. A future study might include whether the defenses on a smaller network somehow transfer to larger networks.

\clearpage
\printbibliography

\end{document}